\documentclass[runningheads]{llncs}
\usepackage[T1]{fontenc}
\usepackage{amsfonts}
\usepackage{booktabs}
\usepackage{dsfont}
\usepackage{multirow}
\usepackage{amssymb}
\usepackage{subfigure}
\usepackage{amsmath, amssymb} 
\usepackage{algorithm}
\usepackage{algpseudocode} 
\usepackage{float}
\usepackage{multicol}
\usepackage{soul}
\usepackage{nicefrac}       
\usepackage{microtype}      
\usepackage{graphicx}
\usepackage{float}
\usepackage{amsmath}
\usepackage{bm}
\usepackage{bbding}
\begin{document}
\title{Zero-Shot Skeleton-Based Action Anticipation}
%
%
\author{Hongsong Wang\inst{1} \and
	Pengbo Yan\inst{2} \and
	Yang Zhang\inst{3} \and
    Qiuxia Lai\inst{4}\textsuperscript{(\Envelope)}
}

\authorrunning{H. Wang et al.}
%
\institute{School of Computer Science and Engineering, Southeast University, Nanjing 210096, China \and
	Southeast University - Monash University Joint Graduate School, SuZhou 215123, China \and
	School of Computer Science and Software Engineering, Shenzhen University, Shenzhen 518060, China \and
	State Key Laboratory of Media Convergence and Communication, Communication University of China, Beijing 100024, China \\
	\email{\{hongsongwang,220255055\}@seu.edu.cn, yangzhang@szu.edu.cn, qxlai@cuc.edu.cn}}
\maketitle              
\begin{abstract}
Action anticipation (AA) aims to recognize ongoing human or humanoids actions from partial observations, enabling robots to predict intentions before the actions are completed. Although skeleton-based AA offers efficiency advantages, existing approaches assume that all action classes are seen during training, which limits their deployment in real-world scenarios where novel actions inevitably arise. To address this gap, we study the new task of Zero-Shot Skeleton-Based Action Anticipation (ZS-SkAA). This task requires recognizing unseen action classes using only limited early-stage skeleton sequences, combining the challenges of partial observations, temporal dynamics, and zero-shot generalization.
To establish foundational research for ZS-SkAA, we introduce:
(1) A baseline model comprising a spatio-temporal feature extractor and a mutual information estimation and maximization module. This baseline model explicitly aligns partial visual features with semantic class embeddings across modalities by estimating and maximizing their mutual information, enhancing generalization to unseen classes.
(2) A benchmark protocol using the NTU RGB+D dataset, which is adapted for rigorous ZS-SkAA evaluation. 
Experiments demonstrate the effectiveness of our model as a strong baseline for ZS-SkAA, achieving high zero-shot accuracy on NTU RGB+D. This work establishes ZS-SkAA as a vital research direction for real-world systems requiring generalization to novel actions.

\end{abstract}
\section{Introduction}
Action recognition and anticipation~\cite{kong2022human,sun2022human,weng2025usdrl,11130651,wang2026data} are fundamental to both digital humans and robots, enabling them to understand and respond to human or humanoid behaviors in real time. In particular, action anticipation, also known as early action recognition, focuses on identifying the category of an ongoing action based on a partial sequence of observations. For instance, if a person bends their knees and lowers their center of gravity, the system should anticipate whether they are about to squat or perform a standing long jump. This capability is essential for time-sensitive applications such as safety monitoring, assistive robotics, and autonomous systems.

\begin{figure}[t]
	\centering
	\label{fig:task}
	\includegraphics[width=\columnwidth]{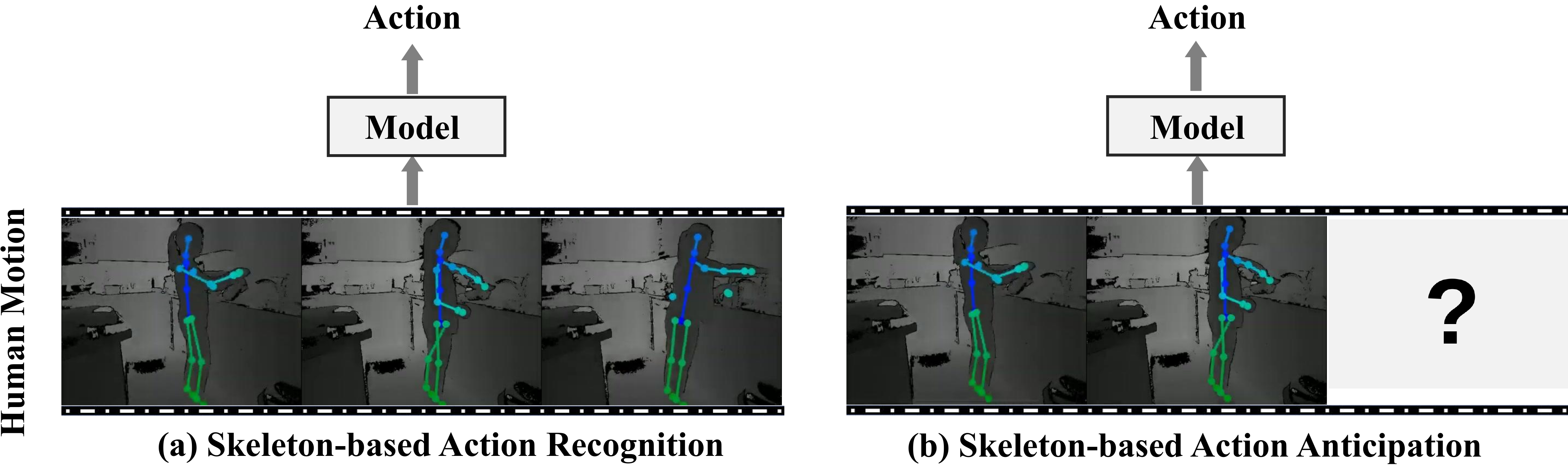} 
	\caption{Comparison of skeleton-based action recognition and action anticipation.}
\end{figure}

Compared to conventional action recognition, action anticipation poses greater challenges. As shown in Figure~\ref{fig:task}, it requires the model to reason under incomplete observations and to distinguish between actions with similar early-stage dynamics. Furthermore, real-world deployment introduces another significant hurdle: many actions encountered during inference may not have been seen during training. Conventional supervised learning methods typically rely on extensive labeled data and fail to generalize effectively in such unseen-class scenarios. Acquiring annotated samples for every possible action class is often impractical or prohibitively expensive, especially in domains like healthcare or surveillance.

To address these limitations, zero-shot learning (ZSL) has emerged as a powerful paradigm that enables models to recognize previously unseen categories by leveraging semantic information shared across tasks~\cite{pourpanah2022review}. While recent advances in ZSL have shown promise in image classification and action recognition, their application to early-stage action anticipation—particularly using skeletal data—remains largely unexplored.

Skeleton-based representations, which encode human motion as structured 3D joint coordinates, have proven especially effective for action understanding due to their robustness to illumination, viewpoint, and background variation. Compared to RGB or depth modalities, skeleton data is lightweight, privacy-preserving, and offers strong generalization across environments. However, existing skeleton-based action anticipation methods assume that all action categories are known at training time~\cite{sadegh2017encouraging,liu2018ssnet,wang2019progressive}, limiting their applicability in open-world settings~\cite{wang2026data}.

In this paper, we address this gap by proposing a new and challenging task: Zero-Shot Skeleton-Based action anticipation (ZS-SkAA). This task combines the difficulty of anticipating future actions from partial observations with the challenge of generalizing to unseen action classes. To the best of our knowledge, this is the first work to formally define and benchmark this task.

To establish a foundation for ZS-SkAA, we introduce a novel baseline model that integrates two key components:
(1) A spatio-temporal feature extraction module that leverages Graph Convolutional Networks (GCNs) for spatial modeling of joint dependencies and Transformer encoders for capturing long-range temporal dynamics.
(2) A mutual information estimation and maximization module that learns a cross-modal alignment between visual skeleton features and semantic class embeddings. 
To further improve predictive performance, we incorporate multi-modal skeletal representations, including Joint, Bone, and Motion, into both modules. This fusion mitigates the limitations of any single modality and enhances the model’s ability to focus on subtle, discriminative motion cues during the early stages of action execution.

Our contributions are summarized as follows:
\begin{itemize}
	\item We introduce Zero-Shot Skeleton-Based action anticipation (ZS-SkAA) as a new task that combines the challenges of action anticipation and zero-shot generalization, addressing a critical gap in real-world action understanding.
	
	\item We propose an effective baseline model that integrates spatio-temporal feature extraction and cross-modal mutual information maximization, supported by a multi-modal skeletal representation, to enhance recognition of unseen actions from partial observations.
	
	\item We establish a benchmark protocol on NTU RGB+D dataset, enabling standardized assessment for future research in this direction.
	
\end{itemize}


\section{Related Work}
\noindent\textbf{Skeleton-Based Action Recognition:}
Skeleton-based action recognition~\cite{11130651,wang2025heterogeneous} has witnessed remarkable progress, primarily driven by the design of powerful spatial-temporal representation models. Early works such as ST-GCN~\cite{yan2018spatial} introduced spatial-temporal graph convolutional networks to effectively capture the dependencies between joints across both spatial and temporal dimensions. This paradigm was further advanced by 2S-AGCN~\cite{shi2019two}, which adaptively learned the graph topology and motion patterns through a dual-stream architecture. InfoGCN~\cite{chi2022infogcn} incorporated the information bottleneck principle to encourage compact and discriminative skeleton representations, while Revisit-ST-GCN~\cite{duan2022revisiting} re-evaluated the design principles behind spatial-temporal GCNs, providing new insights into message-passing and feature aggregation.


However, most existing works on skeleton-based recognition~\cite{chi2022infogcn,shi2019two,yan2018spatial,zhang2021stst,sun2023unified,zhu2023motionbert,duan2022revisiting} are trained under fully supervised settings and assume fixed action categories at test time. In contrast, our work focuses on generalizing to unseen action classes without retraining, which remains under-explored in current skeleton-based models. Our method builds on graph and transformer-based encoders but introduces a zero-shot semantic alignment objective for action anticipation.

\noindent\textbf{Zero-Shot Action Recognition:}
Zero-shot learning (ZSL) aims to recognize previously unseen categories by transferring knowledge from seen classes through an auxiliary semantic space. Classic approaches include attribute-based transfer models~\cite{lampert2009learning}, visual-semantic embedding frameworks such as DeViSE~\cite{frome2013devise}, and cross-modal mapping methods~\cite{socher2013zero}, which project visual features into a shared semantic space (e.g., attributes or word vectors). 
Recent ZSL methods explore mutual information estimation to enhance semantic alignment. Tang et al.~\cite{tang2020zero} and Zhou et al.~\cite{zhou2023zero} leverage mutual information maximization to strengthen the correlation between visual and semantic embeddings. Sylvain et al.~\cite{sylvain2020locality} emphasize locality and compositionality to regularize semantic projections. Specific to skeleton-based action recognition, Jasani et al.~\cite{jasani2019skeleton} proposed a joint pose-language embedding space to enable zero-shot recognition of human actions, bridging skeletal motion patterns and textual semantics.

While zero-shot learning has been extensively studied in image and video domains~\cite{lampert2009learning,socher2013zero,frome2013devise,tang2020zero,fu2015transductive,xian2018feature}, skeleton-based zero-shot action recognition~\cite{jasani2019skeleton,zhou2023zero} remains underexplored. These works typically align entire action sequences to class-level semantics. In contrast, we focus on action anticipation, where only partial skeleton sequences are available, and propose a new method to jointly model motion cues and semantic projections for unseen action anticipation.

\noindent\textbf{Skeleton-Based Action Anticipation:}
Unlike conventional action recognition that assumes full observation, action anticipation targets the classification of ongoing actions given only partial observations. A number of works have addressed this early recognition challenge using skeleton data. SSNet~\cite{liu2018ssnet} introduce a scale selection mechanism to adaptively capture discriminative motion patterns at different observation lengths. DBDNet~\cite{pang2019dbdnet} and PTSL~\cite{wang2019progressive} model bi-directional dynamics and progressive temporal structures respectively, enhancing action anticipation accuracy. Hu et al.~\cite{hu2019early} formulated the task as a soft regression problem, predicting future class probabilities in a continuous manner. To handle uncertain temporal cues and long-term dependencies, Ke et al.~\cite{ke2020learning} propose a latent global network that encodes high-level temporal structure. HARD-Net~\cite{li2020hard} introduces hardness-aware discrimination to focus on challenging temporal segments. Recently, TPA~\cite{stergiou2023wisdom} employs temporal progressive attention guided by consensus among action trajectories to anticipate actions.

Most skeleton-based action anticipation models~\cite{liu2018ssnet,wang2019progressive,pang2019dbdnet,hu2019early,ke2020learning,li2020hard,stergiou2023wisdom} assume access to predefined class labels during training and evaluation. These models do not generalize to unseen classes and rely heavily on temporal supervision. Our work differs by tackling zero-shot skeleton-based action anticipation, where both class-level generalization and partial observation handling are required—a setting not adequately addressed by existing models.


\section{Method}
\begin{figure*}[ht]
	\centering
	\includegraphics[width=1\columnwidth]{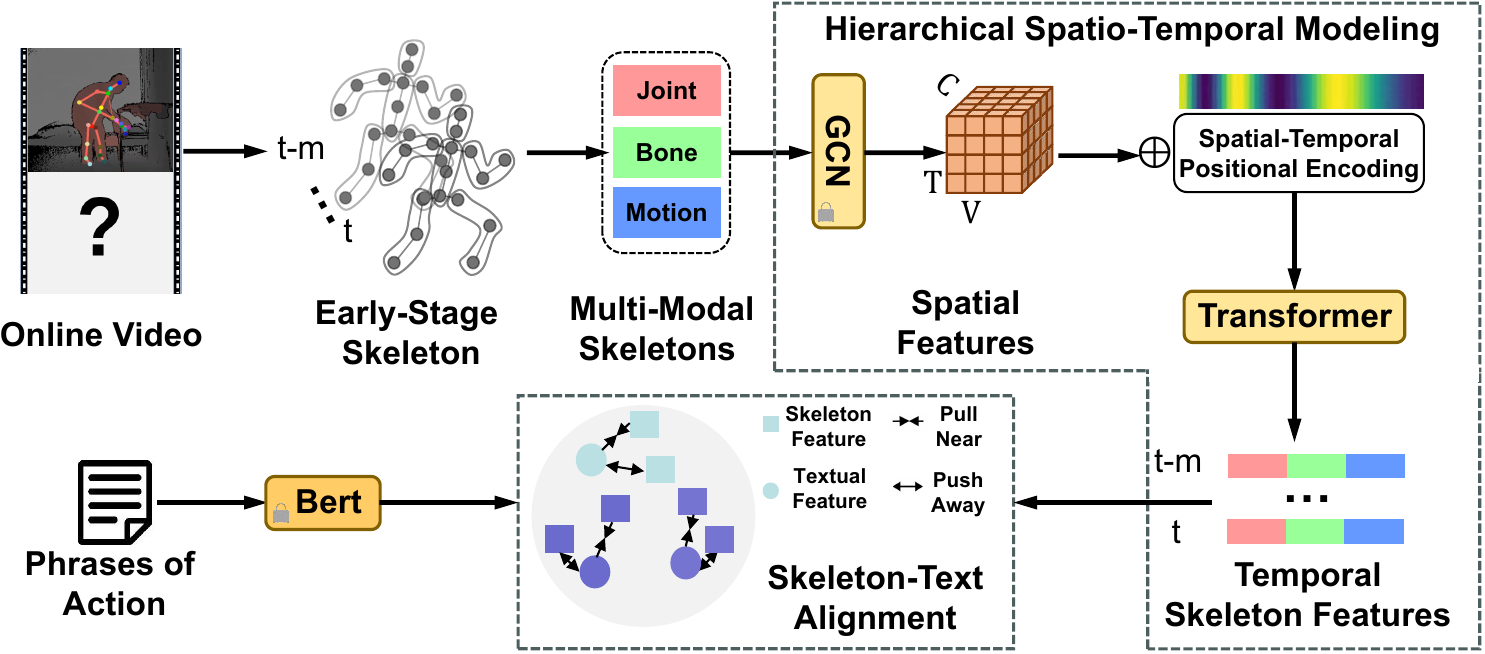} 
	\caption{Pipeline of our approach for zero-shot skeleton-based action anticipation.}
	\label{fig:method}
\end{figure*}

We present a new framework for Zero-Shot Skeleton-Based Action Anticipation (ZS-SkAA), which tackles two key challenges: (1) modeling spatial-temporal dependencies in early-stage skeleton sequences, and (2) aligning visual and semantic representations for unseen class generalization. As illustrated in Figure~\ref{fig:method}, our model comprises three core components: a multi-modal spatio-temporal feature extractor based on GCN and Transformer, a mutual information estimation and maximization module for cross-modal alignment, and a time-aware motion attention mechanism for keyframe localization. 

\subsection{Method Overview}
The challenge of zero-shot skeleton-based action anticipation lies in two main aspects:
(1) insufficient action representation due to partially observed human skeleton sequences;
(2) misalignment between visual features of human skeleton sequences and textual features of action categories.

Given an early-stage skeleton sequence consisting of 3D joint coordinates, our model first encodes spatial and temporal features using a hybrid GCN-Transformer architecture. It then aligns the visual features with semantic class embeddings by estimating and maximizing their mutual information, allowing the model to generalize to unseen action categories. The model further enhances robustness by incorporating joint, bone, and motion modalities and enforcing temporal attention over dynamic keyframes. Finally, action anticipation is performed by computing similarity scores between extracted visual features and candidate semantic embeddings.

\subsection{Hierarchical Spatio-Temporal Feature Extraction}

We adopt a hierarchical architecture that integrates the local spatial modeling capabilities of GCNs with the global temporal modeling strength of Transformers. The input skeleton sequence is a 5D tensor $\bm{X}\!\in\!\mathbb{R}^{N \times C \times T \times V \times M}$, where $N$ is the batch size, $C$ is the feature dimension (e.g., 3 for (x, y, z) coordinates), $T$ is the number of frames, $V$ is the number of joints, and $M$ is the number of persons per frame.
During preprocessing, batch normalization is applied channel-wise to ensure stability across joints and time. To support early-stage action anticipation, the model observes only the first $\rho$ proportion of the sequence, controlled by an early observation ratio $\rho\!\in\!(0, 1]$.

\noindent\textbf{Spatial Modeling with GCN:} Each skeleton frame is naturally represented as a graph, where joints correspond to nodes and bones define the edges connecting them. This structure allows us to apply Graph Convolutional Networks (GCNs) to model the spatial relationships between joints in human skeleton sequences. Specifically, the input tensor of shape $[N,T,V,C]$ is first reshaped to $[N\times V,T,C]$ to facilitate modeling the temporal evolution of each joint independently. A fully connected layer then maps the features into a target dimension before the tensor is reshaped back to $[N,V,T,C]$ for graph processing.
To perform spatial aggregation, the model uses Einstein summation notation to aggregate features from neighboring joints according to a predefined or learnable adjacency matrix. For each joint, the features from connected nodes are aggregated and fused using weighted summation across multiple graph branches. This multi-branch design allows the model to learn relative spatial dependencies from multiple perspectives, significantly improving its ability to perceive complex joint relationships. The GCN module thus serves as a powerful spatial encoder that captures the physical connectivity and structural context of the human body.

\noindent\textbf{Temporal Modeling with Transformer: } After spatial encoding, the output features are fed into a multi-layer Transformer encoder for long-range temporal modeling. 
To compensate for the lack of recurrence, we introduce learnable spatial and temporal positional encodings, allowing the Transformer to distinguish joints and preserve motion order. Specifically, the spatial positional encoding is added to each joint to help the model distinguish anatomically similar joints (e.g., left vs. right wrist), while the temporal positional encoding is shared across all joints in a frame to encode sequence order (e.g., lifting before lowering the arm). Both encodings are trainable and optimized end-to-end, allowing the model to learn discriminative temporal dynamics and contribute to robust action anticipation, especially in early-stage, incomplete motion scenarios.

Overall, the combination of GCN and Transformer fully exploits the spatial and temporal properties of skeleton sequences. While GCN encodes localized joint dependencies, the Transformer captures global temporal evolution, making this hybrid design particularly effective for structured yet low-dimensional inputs like human poses.

\subsection{Mutual Information-Based Skeleton-Text Alignment}

To align partial visual features with semantic class embeddings and enable zero-shot generalization, we adopt a mutual information-based alignment module similar to prior work~\cite{zhou2023zero}. This module consists of two components: a global alignment module for visual-semantic pairing and a temporal constraint module for identifying informative frames.

\noindent\textbf{Global Alignment Module: }
Instead of mapping visual and semantic features to fixed anchor points, we estimate their mutual dependence using a learnable neural similarity function. Specifically, we adopt the Jensen-Shannon-based mutual information estimator~\cite{tschannen2019mutual}, which contrasts matched feature pairs $(\bm{v}, \bm{a})$ with mismatched ones $(\bm{v}', \bm{a})$, optimizes the following objective:
\begin{equation}\label{eq:mutual_info_loss}
	I(V; A) = \mathbb{E}_{(\bm{v}, \bm{a})} [-g(-T(\bm{v}, \bm{a}))] - \\\mathbb{E}_{(\bm{v}', \bm{a})}[g(T(\bm{v}', \bm{a}))],
\end{equation}
where $T(\cdot, \cdot)$ is a trainable scoring function, and $g(\cdot)$ is the $\text{softplus}$ activation function. The mutual information loss $\mathcal{L}_{\text{MI}}\!=\!-I(V; A)$ encourages high similarity for semantically aligned pairs, facilitating knowledge transfer to unseen classes.

\noindent\textbf{Temporal Constraint Module: }
To enhance temporal sensitivity, we incorporate a lightweight keyframe localization strategy based on bidirectional motion attention. For each joint and frame, we compute displacement magnitudes to measure motion intensity:
\begin{equation}
	P_{k,j,c} = (x_{k+1,j,c} - x_{k,j,c})^2 + (x_{k-1,j,c} - x_{k,j,c})^2.
\end{equation}
Frame-wise attention weights are then computed by averaging over joints and channels, followed by normalization. The top-$P$ frames with highest weights are selected as keyframes. 
To encourage the model to focus on these informative moments, we apply a contrastive loss. A masked version of the sequence (with keyframes removed) is passed through the same network, and the drop in mutual information is penalized via:
\begin{equation}\label{eq:loss_time}
	\mathcal{L}_{\text{time}} = \max(0, \beta - (m - m')),
\end{equation}
where $m$ and $m'$ denote mutual information scores of the original and masked sequences, and $\beta$ is a margin hyperparameter. This loss encourages the model to assign higher mutual information to unmasked sequences and implicitly learn to focus on informative frames.

\subsection{Multi-Modal Feature Fusion}

To further enrich spatial-temporal representation, we incorporate a multi-modal input strategy based on joint, bone, and motion modalities. The joint modality $\bm{X}_J\!\in\!\mathbb{R}^{N \times C \times T \times V \times M}$ denotes the original 3D joint coordinates, where $N$ is the batch size, $C$ is the channel dimension, $T$ is the number of frames, $V$ is the number of joints, and $M$ is the number of persons per frame. 
The bone modality is constructed as the relative vector between the joint $\bm{v}$ and its parent node $p(\bm{v})$:
\begin{equation}
	\bm{X}_B(:,:,t,\bm{v},:) = \bm{X}_J(:,:,t,p(\bm{v}),:) - \bm{X}_J(:,:,t,\bm{v},:).
\end{equation}
The motion modality captures the temporal difference between consecutive frames, which is defined as:
\begin{equation}
	\bm{X}_M(:,:,t,\bm{v},:) = \bm{X}_J(:,:,t,\bm{v},:) - \bm{X}_J(:,:,t-1,\bm{v},:).
\end{equation}

After obtaining the bone vector and motion velocity modalities from the joint position modality, we adopt an early fusion strategy to combine the three modalities. The fused data is then fed into the spatio-temporal feature extraction module based on the GCN and Transformer. The subsequent processing follows the same procedure as in the single-modality setting: the resulting visual features and semantic embeddings are jointly input into the mutual information estimation and maximization module for model training and evaluation of its generalization capability on unseen classes.

The final training loss is a combination of mutual information alignment and temporal constraint:
\begin{equation}
	\mathcal{L} = \mathcal{L}_{\text{MI}} + \lambda \mathcal{L}_{\text{time}},
\end{equation}
where $\mathcal{L}_{\text{MI}}$ is the mutual information loss defined in Eq.~(\ref{eq:mutual_info_loss}), $\mathcal{L}_{\text{time}}$ is the temporal constraint loss defined in Eq.~(\ref{eq:loss_time}), and \(\lambda\) balances semantic alignment and temporal robustness. In our experiments, we empirically set $\lambda = 0.5$.

\section{Experiment}

\begin{table}[t]
	\centering
	\caption{Top-1 accuracy (\%) of our model using joint-only and multi-modal inputs under varying early observation ratios. Fusion improves performance in early stages (0.1–0.3), while joint-only models perform better at later stages.}
	\label{tab:table1_results}
	\scalebox{0.95}{
		\begin{tabular}{l|cc|cc}
			\toprule
			\multirow{2}{*}{Early Ratio~~} & \multicolumn{2}{c}{NTU RGB+D 60} & \multicolumn{2}{c}{NTU RGB+D 120} \\
			& ~~Joint~~ & ~~MultiFusion~~ & ~~Joint~~ & ~~MultiFusion~~ \\ \midrule
			0.1 & 39.62 & 40.63 & 26.46 & 28.74 \\
			0.2 & 44.17 & 47.31 & 31.09 & 31.91 \\
			0.3 & 47.39 & 49.35 & 34.16 & 34.64 \\
			0.4 & 50.34 & 50.12 & 36.77 & 36.05 \\
			0.5 & 53.58 & 52.27 & 39.53 & 38.56 \\
			0.6 & 57.59 & 55.02 & 41.60 & 39.78 \\
			0.7 & 59.34 & 57.68 & 42.21 & 41.95 \\
			0.8 & 61.13 & 59.22 & 43.27 & 42.76 \\
			0.9 & 63.64 & 61.72 & 43.49 & 43.81 \\ \bottomrule
	\end{tabular}}
\end{table}

\begin{table*}[t]
	\centering
	\caption{Accuracy (\%) comparison between our method and SMIE~\cite{zhou2023zero} under different early observation ratios on NTU RGB+D 60 and 120. Our method consistently outperforms the baseline, especially in low-ratio scenarios.}
	\label{tab:table2_comparison}
	\begin{tabular}{l|c|cccccccccc}
		\toprule
		\multirow{2}{*}{Dataset} & \multirow{2}{*}{Method} & \multicolumn{9}{c}{Early Observation Ratio} \\
		\cmidrule(lr){3-11}
		&        & ~~0.1~~ & ~~0.2~~ & ~~0.3~~ & ~~0.4~~ & ~~0.5~~ & ~~0.6~~ & ~~0.7~~ & ~~0.8~~ & ~~0.9~~ \\
		\midrule
		\multirow{2}{*}{NTU60} 
		& SMIE~\cite{zhou2023zero}   & 36.50 & 40.48 & 43.47 & 45.31 & 49.16 & 53.88 & 57.13 & 60.82 & 63.58 \\
		& Ours   & 39.62 & 44.17 & 47.39 & 50.34 & 53.58 & 57.59 & 59.34 & 61.13 & 63.64 \\
		\midrule
		\multirow{2}{*}{NTU120}
		& SMIE~\cite{zhou2023zero}   & 23.38 & 25.15 & 27.82 & 29.38 & 31.75 & 34.65 & 38.24 & 42.33 & 43.26 \\
		& Ours   & 26.46 & 31.09 & 34.16 & 36.77 & 39.53 & 41.60 & 42.21 & 43.27 & 43.49 \\
		\bottomrule
	\end{tabular}
\end{table*}

\subsection{Experimental Setup}

\noindent\textbf{Datasets:} We evaluate our method on two large-scale public skeleton-based action datasets: NTU RGB+D 60 and NTU RGB+D 120. 

\textbf{NTU RGB+D 60}~\cite{shahroudy2016ntu} contains 56,578 skeleton sequences from 60 action classes performed by 40 subjects, captured by Microsoft Kinect with 25 joints per person. We adopt the official settings: 1) Cross-Subject (xsub), where half of the subjects are for training, the rest for testing; and 2) Cross-View (view), where training and testing sequences are captured from different viewpoints.

\textbf{NTU RGB+D 120}~\cite{liu2019ntu120} is an extended version with 113,945 sequences across 120 classes performed by 106 subjects. It also provides two splits: 1) Cross-Subject (xsub), where 53 subjects are used for training, the remaining for testing; and 2) Cross-Setup (xset), where sequences captured by even-numbered cameras are used for training, and odd-numbered ones for testing.

\noindent\textbf{Implementation Details: } Our framework employs a two-stage visual encoder: a Graph Convolutional Network (GCN) with 16-channel intermediate features, followed by a Transformer layer, yielding 256-dimensional visual embeddings. For semantic representation, we leverage Sentence-BERT to extract 768-dimensional text features, which are normalized for training stability.
The model is optimized end-to-end using Adam with cosine annealing, trained for 100 epochs with a batch size of 128. To account for dataset scale differences, we set the initial learning rate to 0.00001 for NTU-60 and 0.0001 for NTU-120. A linear warm-up strategy is applied for the first 15 epochs to stabilize early training before transitioning to cosine decay.
We adhere to standardized splits: NTU-60 evaluates on 55 seen and 5 unseen classes, while NTU-120 uses 110 seen and 10 unseen classes, ensuring fair comparison with prior work.

\subsection{Performance Evaluation}
\noindent\textbf{Accuracy under Varying Early Observation Ratios: }
We evaluate the model across early observation ratios from 0.1 to 0.9 under both single-modality (joint only) and multi-modality (joint, bone, motion fusion) settings. As shown in Table~\ref{tab:table1_results}, action anticipation accuracy improves consistently with longer observation windows, due to increased access to motion context and visual-semantic alignment. 

In the early stages (ratios 0.1–0.3), multi-modal fusion yields clear advantages, with up to 4\% improvement over joint-only models. This is because single-modality features lack sufficient motion cues early on, whereas multi-modal fusion incorporates complementary information (bone vectors and motion velocities).
However, as the observation ratio increases (0.4–0.9), the joint-only model outperforms fusion, likely because the joint modality alone captures sufficient context, and additional modalities may introduce noise or redundancy. At high ratios (0.8–0.9), the performance gap narrows.
These results indicate that multi-modal fusion is particularly beneficial for early-stage action anticipation, where limited frames challenge reliable recognition.

\noindent\textbf{Comparison with Existing Methods: } 
We compare our model against SMIE~\cite{zhou2023zero}, a recent method for zero-shot skeleton action recognition adapted here for early action anticipation. As shown in Table~\ref{tab:table2_comparison}, our model outperforms the SMIE at nearly all observation ratios, especially in low-ratio settings where generalization to unseen classes is most challenging. 
The improvement highlights our model’s stronger generalization in zero-shot and early-action scenarios.

\begin{figure}[ht]
	\centering
	\includegraphics[width=0.6\columnwidth]{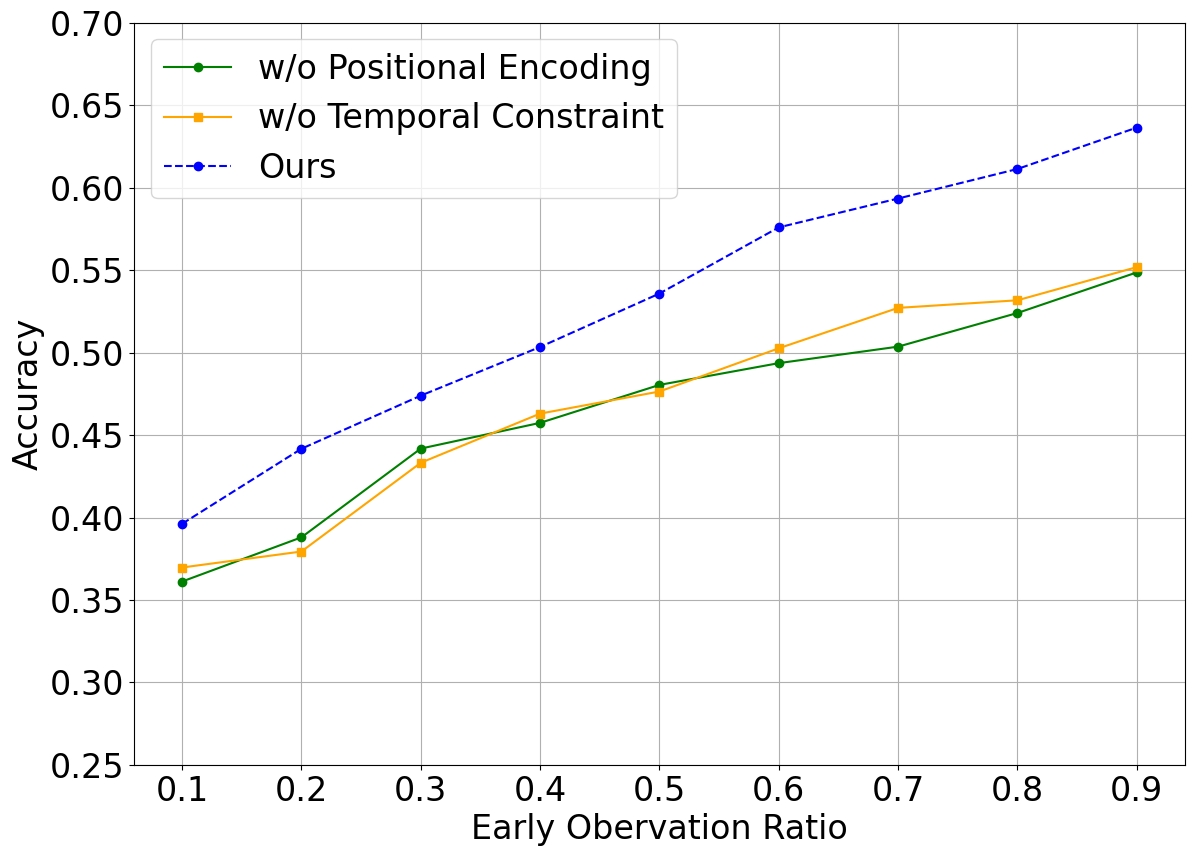} 
	\caption{Ablation studies of our approach on the NTU-60 dataset.}
	\label{fig:ablation}
\end{figure}

\begin{figure}[ht]
	\centering
	\includegraphics[width=\columnwidth]{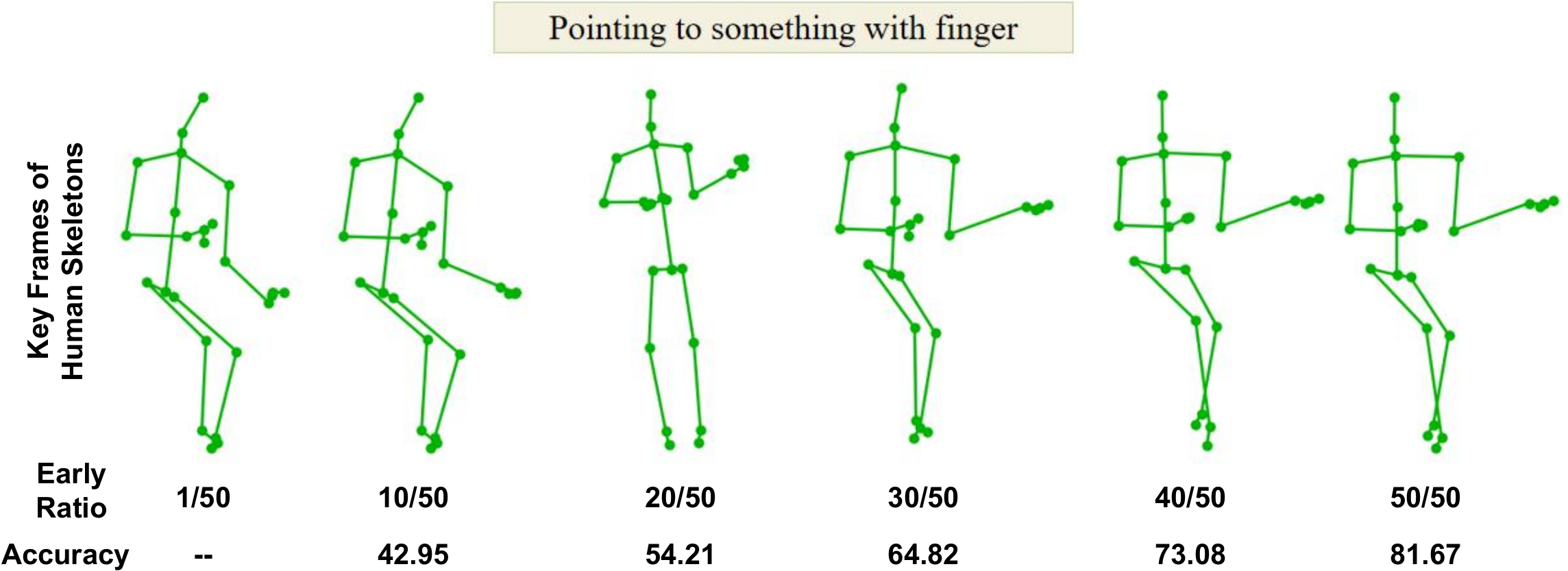} 
	\caption{Visualizations of results of skeleton-based action anticipation.}
	\label{fig:visulizations}
\end{figure}

\subsection{Ablation Studies and Visualizations}
We conduct ablation experiments to evaluate the contributions of two key components: learnable positional encodings and the temporal constraint module. Results on the NTU-60 are shown in Figure~\ref{fig:ablation}.

\noindent\textbf{Impact of Learnable Positional Encoding: }
We evaluate the impact of spatial and temporal positional encodings by removing them from the Transformer and comparing the resulting performance. The absence of positional encoding leads to a noticeable drop in accuracy, particularly at early observation ratios. These results confirm that position-aware encoding is crucial for enhancing feature expressiveness, as it enables the model to distinguish joint identities and preserve temporal order.

\noindent\textbf{Impact of Temporal Constraint Module: }
We also ablate the temporal constraint loss used to encourage focus on discriminative keyframes. Removing this loss results in consistent accuracy degradation. This validates the effectiveness of learning to identify and emphasize dynamic, informative frames during early action anticipation.

\noindent\textbf{Visualization of Skeleton-Based Action Anticipation: }
To better understand the model’s behavior, we visualize skeleton frames at different timestamps for the action ``Pointing to something with finger'' in Figure~\ref{fig:visulizations}, along with action anticipation accuracy at corresponding ratios.
As shown, accuracy improves as the action unfolds, demonstrating how the model refines its predictions with more temporal evidence.


\section{Conclusion}

In this work, we propose Zero-Shot Skeleton-Based Action Anticipation (ZS-SkAA), a new task that addresses the challenge of recognizing unseen action classes from early-stage skeleton sequences. To this end, we introduce a strong baseline model that combines GCN-Transformer-based spatio-temporal encoding with mutual information maximization for cross-modal alignment. We further propose a temporal constraint module to enhance focus on informative frames. Extensive experiments on adapted NTU RGB+D benchmarks demonstrate the model’s effectiveness in zero-shot settings, especially under partial observations. Our results highlight the promise of ZS-SkAA as a practical direction for building generalizable and efficient action understanding systems.

\vspace{0.3cm}
\noindent \textbf{Acknowledgements.} This research was supported by the National Natural Science Foundation of China (Nos. 62302093, 62306292 and 52441503) and the Natural Science Foundation of Jiangsu Province (No. BK20230833).

\vspace{0.1cm}
\noindent \textbf{Disclosure of Interests.} The authors have no competing interests to declare that are relevant to the content of this article.

%
%
%
\bibliographystyle{splncs04}
\bibliography{IEEEfull}

\end{document}